\documentclass{article}

\usepackage{arxiv}

\usepackage{microtype}
\usepackage{graphicx}
\usepackage{subcaption}
\usepackage{booktabs} 
\usepackage[hidelinks]{hyperref}

\usepackage{amsmath}
\usepackage{amssymb}
\usepackage{mathtools}
\usepackage{amsthm}
\usepackage[capitalize,noabbrev]{cleveref}

\theoremstyle{plain}

\theoremstyle{definition}

\theoremstyle{remark}

\usepackage[textsize=tiny]{todonotes}
\usepackage{times}
\usepackage{latexsym}
\graphicspath{ {figures/} }
\usepackage{verbatim}
\usepackage{multirow}
\usepackage{array}
\usepackage{pifont}
\usepackage{xcolor, colortbl}
\usepackage{soul}
\usepackage{algorithm}
\usepackage{float}
\usepackage[T1]{fontenc}
\usepackage[utf8]{inputenc}
\definecolor{LightLightGray}{gray}{0.9}
\usepackage[round]{natbib}
\usepackage{setspace}

\title{Explaining Text Classifiers\\with Counterfactual Representations}

\author{
 Pirmin Lemberger \\
  onepoint\\
  29 rue des Sablons, 75016, Paris (France)\\
  \texttt{p.lemberger@groupeonepoint.com} \\
   \And
 Antoine Saillenfest\\
  onepoint\\
  29 rue des Sablons, 75016, Paris (France) \\
  \texttt{a.saillenfest@groupeonepoint.com} \\
}

\begin{document}
\maketitle
\begin{abstract}
One well motivated explanation method for classifiers leverages counterfactuals which are hypothetical events identical to real observations in all aspects except for one feature. Constructing such counterfactual poses specific challenges for texts, however, as some attribute values may not necessarily align with plausible real-world events. In this paper we propose a simple method for generating counterfactuals by intervening in the space of text representations which bypasses this limitation. We argue that our interventions are minimally disruptive and that they are theoretically sound as they align with counterfactuals as defined in Pearl's causal inference framework. To validate our method, we conducted experiments first on a synthetic dataset and then on a realistic dataset of counterfactuals. This allows for a direct comparison between classifier predictions based on ground truth counterfactuals—obtained through explicit text interventions—and our counterfactuals, derived through interventions in the representation space. Eventually, we study a real world scenario where our counterfactuals can be leveraged both for explaining a classifier and for bias mitigation.
\end{abstract}

\section{Introduction}
\label{section_Intro}
Providing an explanation for the predictions made by a text classifier for a particular document is essential in situations where social bias could have detrimental consequences, for example when documents refer to individuals belonging to different social groups. One well motivated explanation method for classifiers leverages counterfactuals which are hypothetical individuals  identical to real ones in all aspects except for one  feature that is being intervened on \citep{mothilal2020explaining}. Understanding how a classifier reacts to such fictitious individuals will indeed furnish an explanation for how it uses different pieces of information for its predictions. This approach to explanations has been well investigated in the context of social fairness \citep{kusner2017counterfactual, garg2019counterfactual} but it obviously has a wider scope.

Creating counterfactuals (CF) for text documents poses specific challenges when compared to producing CF for tabular data. One of these is related to the fact that it is often by no means obvious how to define a CF text for which the value of some text attribute is modified while everything else is kept unchanged. A number of recent works do, however, propose methods for constructing explicit counterfactuals at the text level in restricted contexts. \citet{zeng2020counterfactual} for instance propose to intervene on entities, separated from their context, to provide CF text samples that can be used for improving the generalization of a NER classifier under limited observational samples. \citet{calderon2022docogen} also intervene on the text by replacing some domain-specific terms to create coherent counterfactuals also used for data augmentation purposes. \citet{madaan2021generate} perform controlled text generation to enforce some user provided label. 

In this paper, leveraging insights from recent research on concept erasure \citep{elazar2021amnesic,belrose2023leace,shao-etal-2023-gold}, we propose a simple method for producing counterfactual representations (CFR) defined as interventions on text representations produced by a generic neural encoder like BERT. A CFR thus implements the alteration of the value of a single protected text attribute. Although the corresponding information is spread over all components of a high-dimensional representation we ensure that our interventions are minimally disruptive in a precise sense. More importantly, CFRs can be instantiated even in cases where a corresponding intervention on the text proves impossible. Finally, these CFRs moreover turn out to be easy to compute. 

\subsection*{Use cases of counterfactuals representations}
\label{subsection_UCRC}
Our method for creating CFRs for texts can be applied to various use cases. It will however prove especially valuable in scenarios where direct interventions on the texts would either lack meaning or incur excessive costs, whether it is human labour or using a generative AI service. Explaining why some texts have been classified in an unexpected category is one important use case. Our method can indeed isolate the role of specific values of a concept in a classifier's prediction whereas traditional erasure methods only provide an evaluation of the  global impact of a concept, typically as an average treatment effect.

The value-by-value analysis we propose will be of particular interest when the fairness of a classifier is at stake because it will reveal precisely which demographic groups are discriminated against. Beyond this explanation use case our CFRs can also serve for counterfactual data augmentation which consists in adding CF to an existing train set. This task is generally performed for OOD generalizability improvement \citep{kaushik2020learning, kaushik2021explaining} or model robustness and fairness \citep{garg2019counterfactual}.

\subsection*{Contributions}
\label{subsection_Contrib}
Our contributions are as follows:
\begin{enumerate}

\item We propose a simple method for generating textual counterfactual representations which corresponds to replacing one concept value with another (sections \ref{subsection_BG} and \ref{subsection_MinimalInterv}).

\item Beyond intuitive arguments, we show that our model aligns with the definition of counterfactuals in Pearl's causal inference framework and is thus theoretically sound (section \ref{subsection_RelatPearl}).

\item We introduce the EEEC+ synthetic dataset, which enables the generation of genuine counterfactuals by performing explicit interventions on texts (section \ref{section_Datasets}) that will serve as a ground truth when we compare the response of a classifier to these with the response to our counterfactuals (sections \ref{sec:direct_comp} and \ref{sec:treatment_effect}).

\item We exhibit practical use cases for our counterfactuals in realistic contexts where counterfactuals are generally not available. We use them to evaluate the causal effects in a sentiment prediction task (section \ref{sec:treatment_effect_realistic}) and to explain a classifier's prediction (section \ref{sec:explaining}).

\end{enumerate}

The aim of this work is not to achieve any kind of SOTA performance. Instead, we aim to demonstrate the usefulness of a simple and theoretically sound regression-based approach to generate counterfactual representations, which can serve as a strong baseline for a variety of tasks. Code and data are available on github\footnote{\href{https://github.com/ToineSayan/counterfactual-representations-for-explanation}{github.com/ToineSayan/counterfactual-representations-for-explanation}}.

\section{Related Work}
\label{section_RW}
Using Pearl's causal inference framework for defining counterfactual fairness was pioneered in \citet{kusner2017counterfactual}. This work motivates and formalizes the intuition that a classifier which is fair towards individuals belonging to different social groups should produce the same predictions for an actual individual and for a counterfactual individual belonging to a different group, other things being equal. It also stresses the importance of taking into account causal relationships between the variables that describe an individual when constructing fair classifiers. These relations are typically expressed with a DAG associated to the structural causal model (SCM) \citep{pearl2009causality, peters2017elements} which describes the data generation process. One central observation is that fair predictors should only rely on variables that are non-descendant of protected variables in the causal graph. Finally, the authors describe an algorithm for training fair classifiers that uses a deconvolution approach. Our method for producing CFR for text representations could be used as a practical way of identifying which values of a sensitive text attribute imply a violation of the counterfactual fairness of a possibly biased classifier.

A slightly stronger notion of counterfactual invariance (CFI) is introduced in \citet{veitch2021counterfactual} in order to formalize what it means for a classifier $\widehat{Y}$ to successfully pass stress tests which involve intervening on a protected attribute $Z$. Intuitively, a CFI classifier $\widehat{Y}$ does not rely on that part of the information in $X$ that can be causally affected by the value of $Z$. The main result of this work is that, depending on the underlying causal structure of the data generating process, a CFI predictor $\widehat{Y}$ obeys different independence relations that form a testable signature of the desired invariance. Our method for creating CFR also relies on a part $X^\perp$ that is unrelated to a protected attribute $Z$, although in a weaker sense. But, unlike \citet{veitch2021counterfactual} and following \citet{shao-etal-2023-gold}, we exhibit $X^\perp$ explicitly under some linearity assumptions on how the sensitive information $Z$ is hidden in $X^\perp$.

Another line of work 
\citep{
ravfogel-etal-2020-null,
xie2017controllable,
elazar2021amnesic, 
belrose2023leace, 
feder2021causalm},
focuses on defining methods for concept erasure. The aim is still to build fair predictors that use data from which information on protected attributes has been ‘‘scrubbed''. However, achieving such erasure can be tricky due to the presence in the text of numerous factors correlated with the concept to be erased \citep{de2019bias}. To circumvent this difficulty, one possibility is to intervene on text representations rather than on texts themselves \citep{barr2021counterfactual}. Intervention methods on representations generally fall into two categories: adversarial methods and linear methods. The former rely on a gradient-reversal layer during training to produce representations that do not encode information about the protected attribute \citep{feder2021causalm}, but have been proven to fail at fully removing this information \citep{elazar2018adversarial}. Focusing on linear methods, \citep{ravfogel-etal-2020-null,elazar2021amnesic,belrose2023leace,shao-etal-2023-gold} use projections that remove unwanted information from the representation space. In our work we also opt to intervene on representations using a closed form projector acting on the representation space as in \citet{belrose2023leace}. Two aspects of this approach are worth mentioning. First, there is no need to train a machine learning model, and therefore it requires minimal computational resources. Second, except for the erased information, this method preserves as much information as possible in a precise sense.

Finally, several papers propose counterfactual benchmarks. In \citet{de2019bias}, approximate CFs are generated to assess the impact of gender information on the occupation classification in HR systems. \citet{abraham2022cebab} proposes a benchmark for explanation methods consisting of a large set of interventions on short restaurant reviews. \citet{feder2021causalm} evaluates the causal effect of a concept on a classification task using synthetic counterfactuals. In most cases the interventions are defined for binary attributes only. As our method can be applied beyond scenarios with binary attributes, we introduce in this study a counterfactual benchmark dataset with a non-binary attribute. 

\section{Creating Counterfactual Representations}
\label{section_CreateCF}

\subsection{Background}
\label{subsection_BG}
Our aim is to define CFRs that can be used as reliable substitutes for genuine CF. Let us thus start by enumerating what we intuitively expect from a ``good'' CFR. For the sake of clarity, let's assume that a sentence $s$ describes the emotional state of an individual and that it is represented by an embedding $X(s)\in\mathbb{R}^d$, obtained from a standard encoder like BERT. Suppose that $\widehat{Y}$ is a classifier for some discrete $Y$ like the emotional content conveyed by $s$, and that $Z(s)$ is a discrete protected attribute like the gender or the race of the individual referred to in $s$. We thus assume the causal graph is $Z\to X \to \widehat{Y}$.\footnote{The causal relationships with $Y$ do not concern us because we focus on explaining predictions $\widehat{Y}$.} 

Now suppose that starting from a text $s$ conveying a race $Z(s)$ we can explicitly exhibit a CF text $s_{Z\leftarrow z}$ referring to a hypothetical individual whose race is $Z(s_{Z\leftarrow z})=z\neq Z(s)$, all other things being equal. Let $X(s_{Z\leftarrow z})$ be the representation of this CF sentence and let $X(s)_{Z\leftarrow z}$ be a tentative CFR obtained by intervening directly on $X(s)$. From a ``good'' CFR $X_{Z\leftarrow z}$ we expect that:
\begin{itemize}

\item it should fool any classifier $\widehat{Y}$ most of the time, namely we expect that $P[\widehat{Y}(X(s_{Z\leftarrow z}))\neq \widehat{Y}(X(s)_{Z\leftarrow z})]$ is small in a sense to be made precise,

\item similarly, it should fool any classifier $\widehat{Z}$ most of the time, 

\item the CFR $X_{Z\leftarrow z}$ should preserve as much information in $X$ as possible, except for that part on which we intervene to change the value of $Z$,

\item finally, calculating $X_{Z\leftarrow z}$ from $X$ should be computationally inexpensive.
\end{itemize}

\subsection{Making Minimal Interventions}
\label{subsection_MinimalInterv}
To define a CFR as a minimally disruptive intervention we follow
\citet{ravfogel2023log} which introduced the concept of linear guardedness that we now briefly review. It formalizes the intuition that only part of the information in $X$ is useful for predicting $Z$ with a linear predictor. Let thus $\eta(X)$ be a predictor for $Z$ and assume that the loss function $\ell(\eta,Z)$ is convex in its first argument, which is the case for the usual cross-entropy for instance. A representation $X^\perp$ is then said to linearly guard $Z$ (as a one-hot encoded variable for $k$ categories in $\{0,1\}^k$) if no linear predictor $\eta(X^\perp)=\mathbf{W}X^\perp+\mathbf{b}$ is able to predict $Z$ better than a constant predictor $\eta(X^\perp)\equiv\mathbf{b}$. More formally, $X^\perp$ as a function of the text $s$ should maximize the minimum expected loss over linear predictors:
\begin{equation}
	X^\perp\in\max_X\min_{\mathbf{W}, \mathbf{b}} \mathbb{E}[\ell(\eta(X),Z)].
\end{equation}
The linearity assumption is a strong one\footnote{Non-linear concept erasure is still largely an open problem. Three lines of work have tackled it recently. Adversarial approaches \citep{feder2021causalm}, kernelized versions of linear erasure methods \citep{ravfogel2022adversarial} and more recently approaches leveraging rate distortion theory \citep{basu2024robust}.} but this is the price to pay for having the useful equivalence in \citep{belrose2023leace} that $X^\perp$ linearly guards $Z$ iff 
\begin{equation}
	\boldsymbol{\Sigma}_{X^\perp Z}:=\mathrm{Cov}[X^\perp,Z]=0.
\end{equation}
Moreover \cite{belrose2023leace} show that such a protected $X^\perp$ can be obtained from an arbitrary representation $X$ by a simple projection 
\begin{equation}
	X^\perp = \mathbf{P}X
\end{equation}
provided $\mathbf{P}$ satisfies $\mathrm{ker}(\mathbf{P})\supseteq\mathrm{im}(\boldsymbol{\Sigma}_{XZ})$. In words, the projector $\mathbf{P}$ should nullify the column space of the covariance matrix $\mathbf{\Sigma}_{XZ}:=\mathrm{Cov}[X,Z]$. In general $\mathbf{P}\neq\mathbf{P}^\top$ and thus the projection is oblique and is not unique.\footnote{In \citet{belrose2023leace} this freedom is used to make $X^\perp=\mathbf{P}X$ as close as possible to the original $X$ in quadratic mean.} We simply use the orthogonal projector on $\mathrm{im}(\mathbf{\Sigma}_{XZ})^\perp$. Denoting $V^\parallel=\mathrm{im}(\boldsymbol{\Sigma}_{XZ})$, which has dimension $k-1$ (as $Z$ is one-hot), and $V^\perp$ its orthogonal complement in $\mathbb{R}^d$, we decompose the representation space $\mathbb{R}^d$ as $V^\perp\oplus V^\parallel$. 

\begin{figure}[t]
    \centering
    \includegraphics[width=0.7\textwidth]{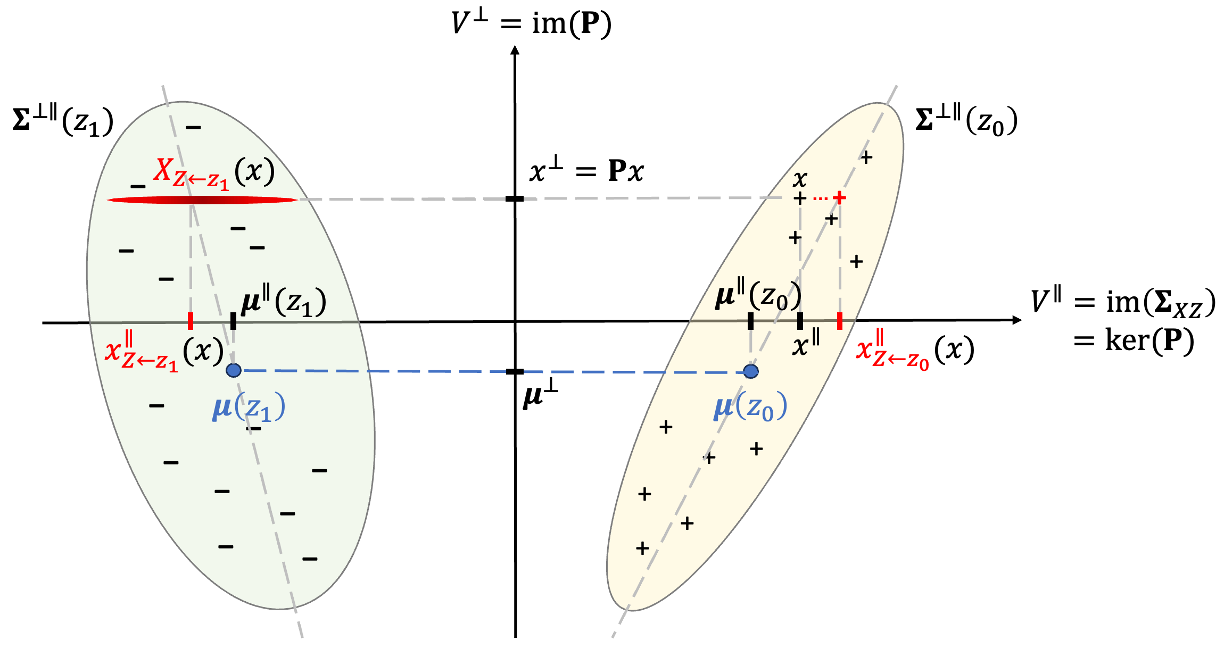}
    \caption{The representation space when $Z$ takes $k=2$ values. Representations of texts for which $Z(s)=z_0$ are shown as $+$ and those for which $Z(s)=z_1$ as $-$, they form two clusters. The representation $x$ is associated with a text for which $Z=z_0$. Once projected by $\mathbf{P}$ on $V^\perp$ we obtain a representation $x^\perp$ from which it is impossible to recover the value $z$ of the protected attribute $Z$ using a linear predictor. This information is contained in $x^\parallel$. Our CFRs $x_{Z\leftarrow z_0}$ and $x_{Z\leftarrow z_1}$ for $x$ corresponding to setting $Z=z_0$ or $z_1$ are obtained by regressing $x^\parallel$ on $x^\perp$ on observations for which $Z=z_0$ and $z_1$ respectively (oblique dashed lines). The random variable $X_{Z\leftarrow z_1}(x)$ is the $Z=z_1$ non deterministic Pearl counterfactual for $x$. Its expectation value corresponds to our CFR $x_{Z\leftarrow z_1}(x)$. The remaining notations are defined in equations (\ref{eq_X_Z}), (\ref{eq_reglin2}) and (\ref{eq_SCM}).
}
    \label{fig_projection}
    \vspace{1em}
\end{figure}

We typically have $k\ll d$ which means that $\mathbf{P}$ erases only a tiny fraction of the information in $X$, namely that information which could be used to predict $Z$ using a linear predictor. Figure \ref{fig_projection} illustrates the geometric situation when $Z$ can take $k=2$ values. In particular the component $x^\parallel$ contains information allowing to predict $Z$ (linearly) whereas the component $x^\perp$ does not.
\begin{figure}[ht]
    \centering
    \includegraphics[width=0.25\textwidth]{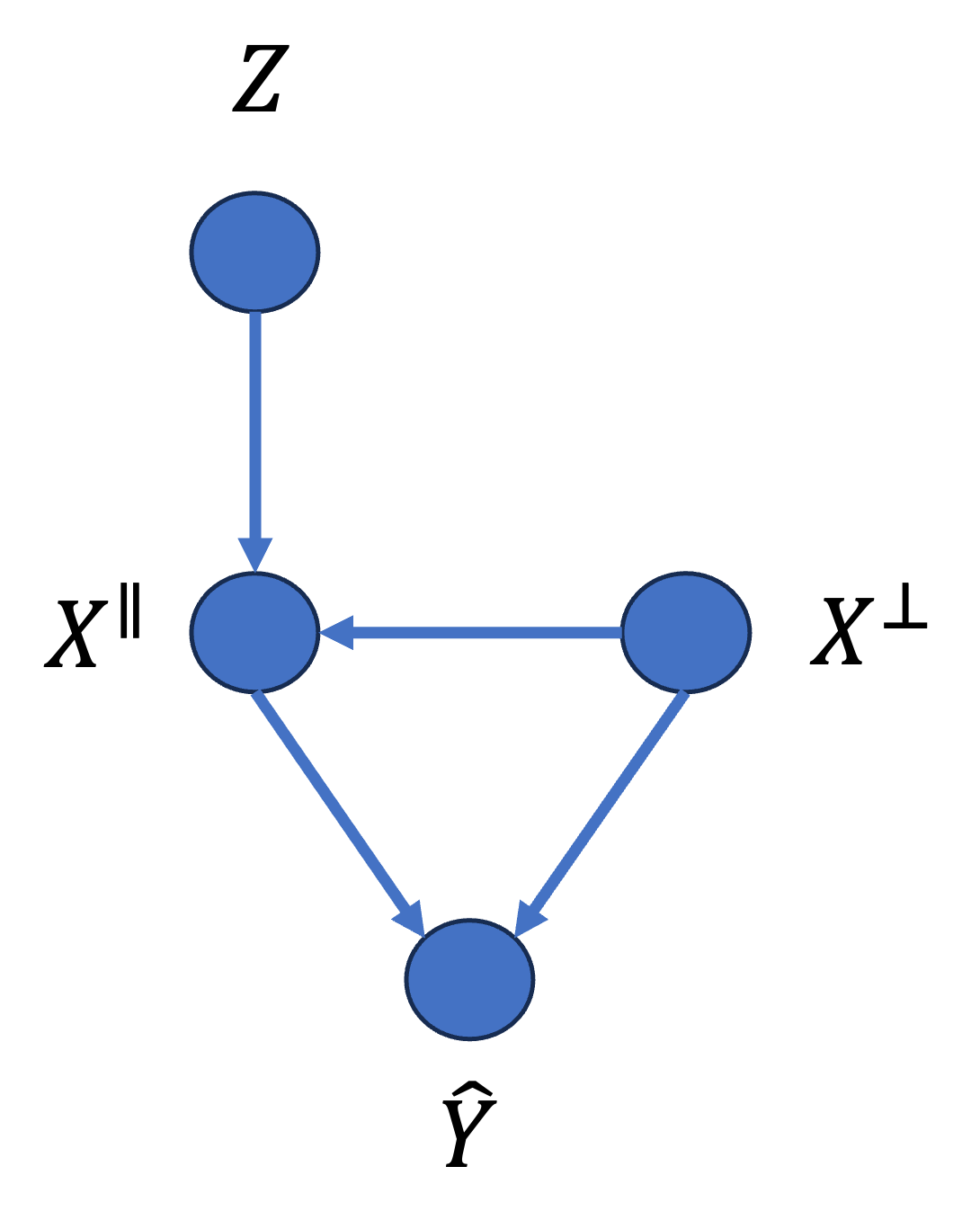}
    \caption{ The DAG $G$ which corresponds to the SCM generating model of text documents.}
    \label{fig_SCM}
    \vspace{1.5em}
\end{figure}

To define our CFR $x_{Z\leftarrow z}$ for an initial representation $x$ we first set $x^\perp_{Z\leftarrow z} := x^\perp$, thus keeping the component without linear information on $Z$ unchanged. Next, we define $x^\parallel_{Z\leftarrow z}$ by linearly regressing $x^\parallel$ on $x^\perp$, on the subset of texts for which $Z(s)=z$. The CFR $x_{Z\leftarrow z}$ will thus be close (in quadratic mean) to representations of real sentences $s$ having $Z(s)=z$ as illustrated in Figure \ref{fig_projection}. More precisely, we define $x_{Z\leftarrow z}^\parallel$ using one multivalued least square regression for each possible target value $z$. Summarizing, our CFR's are thus defined by
\begin{equation}
\label{eq_reglin1}
    x_{{Z\leftarrow z}}(x) := 
\begin{bmatrix}
     x^\parallel_{{Z\leftarrow z}}(x)
     \\
\vspace{-5pt} 
\\
     x^\perp_{{Z\leftarrow z}}(x)
\end{bmatrix}:=
\begin{bmatrix}
     \mathbf{W}(z)\:x^\perp(x) + \mathbf{b}(z)
     \\
\vspace{-5pt} 
\\
     \!x^\perp(x)
\end{bmatrix}.
\end{equation}
\subsection{Relation with Pearl's Framework}
\label{subsection_RelatPearl}
In this section we will argue that the CFR $x_{Z\leftarrow z}$ defined above fits naturally into Pearl's causal inference framework. For this we will exhibit an appropriate structural causal model (SCM) the definition of which we now briefly recall to fix the notations, referring to 
\citep{pearl2009causality,peters2017elements} for more details.

A SCM specifies the causal mechanism that generates data. It is defined by a DAG $G$ where each node $i\in G$ is associated to an observed variable $O_i$ and a noise variable $U_i$. A set $\{f_1,\dots,f_{|G|}\}$ of functions specifies how each variable $O_i$ depends on its parent variables $\mathrm{PA}_i$ in $G$ and on the noise variable $U_i$, namely $O_i=f_i(\mathrm{PA}_i\,;U_i)$. The set of observed and noise variables are denoted by $O$ and $U$ respectively. At last, let $P(U)$ denote the joint probability distribution of the noise variables $U$, which are assumed independent and which are the only source of randomness for the observed variables $O$. Let $P(O)$ denote the induced probability distribution. An intervention on a variable $Z=O_i\in O$ which sets its value to $z$ is defined by replacing $f_i(\mathrm{PA}_i\,;U_i)$ by the constant function $f_i\equiv z$. If $X\in O$ is another variable of interest, such an intervention induces a modified distribution on $X$ which we denote by $P(X_{Z\leftarrow z})$. If we observe that some variables $E\subset O$ have values $e$ this induces a conditional distribution $P(U|E=e)$ on the noise variables (which need not be independent anymore). What would have been the value of the variable $X$ if the value of $Z$ had been equal to $z$? The answer to this counterfactual question is given by first conditioning the noise variables $U$ on the evidence $E=e$ (abduction) and then by an intervention which sets $Z=z$ on this modified SCM. More precisely, the counterfactual distribution for $X$ is defined by $P(X_{Z\leftarrow z}|E=e)$.

In our case $O=\{Z, X^\perp, X^\parallel, \widehat{Y}\}$. The DAG $G$ of the relevant SCM is shown in Figure \ref{fig_SCM}. It expresses the fact that the prediction $\widehat{Y}$ depends on the variable $Z$ only through $X^\parallel$ and that $X^\parallel$ and $X^\perp$ are correlated as revealed by experiment and displayed in Figure \ref{fig_projection}.

The protected variable we act upon is $Z$. Let us assume it is a balanced categorical variable with $k$ values $Z\sim \mathrm{Cat}(\frac{1}{k},\cdots,\frac{1}{k})$. Let us moreover assume that $X=(X^\perp, X^\parallel)$ is distributed as a multivariate Gaussian whose mean $\boldsymbol{\mu}(z)$ and covariance $\boldsymbol{\Sigma}(z)$ depend on $z$. The conditional distribution $P(X|Z=z)$ is thus given by
\begin{eqnarray}
\nonumber
	P(X|z)\: = \mathcal{N}(\boldsymbol{\mu}(z), \boldsymbol{\Sigma}(z)), 
	\\ \label{eq_X_Z}
    \boldsymbol{\mu}(z) :=
	\left[
	\begin{array}{l}
    \!\!\boldsymbol{\mu}^\perp \!\! \\
    \!\!\boldsymbol{\mu}^\parallel(z) \!\!
	\end{array}
\right],
\:\:
	\boldsymbol{\Sigma}(z) :=    
    \begin{bmatrix}
    \!\!\!\!\!\!
    \boldsymbol{\Sigma}^{\perp\perp} & 
    \boldsymbol{\Sigma}^{\perp\parallel}(z)  \\
    \boldsymbol{\Sigma}^{\parallel\perp}(z) & 
    \boldsymbol{\Sigma}^{\parallel\parallel}(z)
    \end{bmatrix},
\end{eqnarray}
where both $\boldsymbol{\mu}^\perp$ and $\boldsymbol{\Sigma}^{\perp\perp}$ are independent of $z$ because $X^\perp$ is not impacted by $z$. Using standard properties of multivariate Gaussian we infer from (\ref{eq_X_Z}) that the conditional $P(X^\parallel|X^\perp=x^\perp, Z=z)$ is linear-Gaussian for each $z$
\begin{eqnarray}
\nonumber
	P(X^\parallel| x^\perp, z) \:=\:
\mathcal{N}\left(\boldsymbol{\mu}^\parallel(x^\perp,z),\boldsymbol{\Sigma}^\parallel(z)\right),\\
	\label{eq_reglin2}
	\boldsymbol{\mu}^\parallel(x^\perp,z):=\mathbf{W}(z)\:x^\perp + \mathbf{b}(z),
\end{eqnarray}
where $\boldsymbol{\Sigma}^\parallel(z), \mathbf{W}(z), \mathbf{b}(z)$ can be expressed as closed-form expressions involving the components of $\boldsymbol{\mu}(z)$ and $\boldsymbol{\Sigma}(z)$. An SCM which is compatible with the above can now be proposed by introducing appropriate noise variables $U_Z, U^\perp, U^\parallel$ and linear functions $f_Z, f_{X^\perp}, f_{X^\parallel}$ associated with the $Z, X^\perp, X^\parallel$ nodes in $G$. The predictor $\widehat{Y}$ is identified with $f_{\widehat{Y}}$ and has no associated noise variable. As we shall argue below, the noise variable $U^\parallel:=(U^\parallel_1,\dots,U^\parallel_k)$ for $X^\parallel$ should have as many components as the number of values $Z$ can take which is $k$. We shall write $U^\parallel(Z)$ to mean $U^\parallel_z$ when $Z=z$. Using the definitions for $\boldsymbol{\mu}^\perp, \boldsymbol{\mu}^\parallel(x^\perp,z)$ and $\boldsymbol{\Sigma}^{\perp\perp}$ introduced in (\ref{eq_X_Z}) and (\ref{eq_reglin2}) we then define the SCM which implements the distribution $P(O)$ and the causality relations defined by $G$ as
\begin{align}
\nonumber
Z & =f_Z({U_Z}):= U_Z, 
&U_Z 
\sim&\: \mathrm{Cat}(\textstyle{\frac{1}{k},\dots,\frac{1}{k}}),
\\
\nonumber
X^\perp &= f_{X^\perp}(U^\perp) := \boldsymbol{\mu}^{\perp} + \mathbf{\Sigma}^{\perp\perp} U^\perp, 
&U^\perp 
\sim&\: \mathcal{N}(\mathbf{0},\mathbf{1}),
\\
\nonumber
X^\parallel &= f_{X^\parallel}(X^\perp,Z\,; U^\parallel) 
& U^\parallel_z 
\sim &\: \mathcal{N}(\mathbf{0},\mathbf{1}) \text{ for } z =1,\dots,k,
\\ 
\nonumber & := \boldsymbol{\mu}^\parallel(X^\perp,Z)
+ 
\boldsymbol{\Sigma}^\parallel(Z)U^\parallel(Z) & &
\\
\label{eq_SCM}
\widehat{Y} &= f_{\widehat{Y}}(X^\perp,X^\parallel).
\end{align} 
Suppose now that we observe the evidence $e:=(Z(s), X^\perp(s), X^\parallel(s)):=(z, x^\perp, x^\parallel)$ for some text $s$. Abduction amounts to reading off the values $u^\perp$ and $(u^\parallel_1,\dots,u^\parallel_k)$ of the noise variables from (\ref{eq_SCM}). Once conditioned on the evidence, $U^\parallel_z$ and $U^\perp$ are obviously not random anymore, so neither are $Z$ and $X^\perp$. The $X^\parallel$ variable on the other hand remains stochastic because knowing that $Z=z$ only freezes $U^\parallel_z$ but not $U^\parallel_{z_1}$ for $z_1\neq z$. Next, the counterfactual distribution $P(X_{Z\leftarrow {z_1}}|E=e)$ is defined by acting on $Z$ to set its value to $z_1\neq z$ in the SCM (\ref{eq_SCM}) in which the only remaining noise is $U^\parallel_{z_1}$. The distribution $P(X^\parallel_{Z\leftarrow z_1}|E=e)$ is thus given by (\ref{eq_reglin2}) by replacing $z$ by $z_1$. Its expectation $\boldsymbol{\mu}^\parallel(x^\perp,z)$ is nothing but the $\parallel$ component of our CFR $x_{Z\leftarrow z}(x)$ as defined by a linear regression in (\ref{eq_reglin1}). The $\perp$ component is deterministic $X^\perp_{Z\leftarrow z}=x^\perp$ in agreement with the second component in (\ref{eq_reglin1}). In other words, our CFR is nothing but the expectation of a counterfactual as defined in Pearl's causal inference framework for the SCM (\ref{eq_SCM}), thus justifying our claim.

If we had refrained from defining as many noise variables $U^\parallel_z$ as there are different values of $z$, then the evidence $e$ would have fully determined the value of $U^\parallel$, making the CFR fully deterministic. However, this would induce a geometric relationship between the locations of a representation $x$ and that of its counterfactual $x_{Z\leftarrow z}$ for which there is no justification whatsoever in any text generation mechanism.

\section{Datasets and training details}
\label{section_Expe}
To assess our CFR generation model, we conducted a series of experiments on both synthetic and real world datasets. 
We first introduce a synthetic dataset, named EEEC+, to provide a ground truth in the form of CFs defined at the text level to which we later compare our CFRs. Next, we test CFRs for assessing causal effects on the realistic benchmark dataset CEBaB \citep{abraham2022cebab}, which poses a greater challenge than EEEC+ because it is much smaller, sparse in concept labels and in which concept values are not determined by a local signal in the samples. At last, we leverage the real world dataset BiasInBios \citep{de2019bias} to challenge our CFRs to provide useful substitutes for CFs in cases these are not available. This will motivate the practical usefulness of our CFRs as a tool for classifier explainability. 

\subsection{Datasets}
\label{section_Datasets}

\paragraph{EEEC+} We introduce a new synthetic dataset, EEEC+, as an extension of the existing EEEC dataset \citep{feder2021causalm}. Both are well suited for evaluating the impact of protected attributes (the gender or perceived race of the individual referred to in a text) on downstream mood state classification. Compared with EEEC, besides increasing the diversity of templates, we also turned the binary race concept into a ternary one to extend the scope of evaluation of our CFR model. Information on the creation and structure of EEEC+ can be found in Supplementary Material \ref{sec:EEEC+}. 

Each observation in EEEC+ is labelled with a binary gender (male of female), a ternary race (white American, Afro-American or Asian-American, which incidentally allows to go beyond just flipping a pair of races) and a mood state (joy, fear, sadness, anger or neutral). 

We built both a balanced and an aggressive version of EEEC+. In the balanced version, mood state is uncorrelated with gender or race. In the aggressive version, a correlation has been induced by assigning 80\% of 'joy' states and 20\% of other mood states one specific value of the protected attribute (female for gender or Afro-American for race). Each observation in the balanced version of EEEC+ has been assigned one genuine CF generated by randomly selecting a counterfactual value for the protected attribute and automatically editing the text accordingly. Every EEEC+ version comprises 40,000 observations distributed across three stratified-by-mood-states splits, with 26,000 training (65\%), 6,000 validation (15\%), and 8,000 test samples (20\%). 

\noindent \paragraph{CEBaB} This realistic dataset is well suited for evaluating the causal effect of a concept on a sentiment classification task \citep{abraham2022cebab}. It includes both 2,299 original restaurant reviews from OpenTable and human-edited counterfactual reviews in which an aspect of the dining experience (food, service, ambiance or noise) was modified. The analysis of causal effects in CEBaB is facilitated through the creation of edit pairs. These are pairs of observations from the same edit set that differ in the value of one aspect. An edit set comprises an original observation and all observations edited from that original observation. Observations have been annotated with multiply-validated sentiment ratings at the aspect level (mostly Positive, Negative, or Unknown labels) and at the review level (1 star (very negative) to 5 stars (very positive)). In this article, we use CEBaB's exclusive train set described in \citet{abraham2022cebab} as training data. So we conducted the study on 5,117 observations distributed across three splits, with 1,755 training (34\%), 1,673 validation (32\%), and 1,689 test samples (33\%).

\noindent\paragraph{BiasInBios} This real world dataset is suited for studying gender biases in biography classification tasks \citep{de2019bias}. It consists of short biographies collected through web scraping and labeled with binary gender and occupation (28 occupations in total). This dataset is notoriously gender-biased. We have used the dataset version introduced in \citet{ravfogel-etal-2020-null} which contains over 98\% of the original dataset, as the full version is no longer available on the web. It comprises 399,423 biographies distributed across three stratified-by-occupation splits, with 255,710 training (65\%), 39,369 validation (10\%), and 98,344 test samples (25\%). 

\subsection{Training details}
\label{section_Traning_Details}

In the subsequent analysis, each genuine observation is represented by the last hidden state of a frozen BERT (bert-base-uncased) \citep{devlin2019BERT} over the [CLS] token. The component $X^{\perp}$ results from an orthogonal projection onto $V^{\perp}$. The computation of $\boldsymbol{\mu}^\parallel(x^\perp, z)$ results from a linear regression via stochastic gradient descent with mean squared error objective and $L^2$-regularization for each value $z$ of $Z$. Unless specified, we use the deterministic version of $x^\parallel_{Z\leftarrow z}(x^{\perp})$ defined in (\ref{eq_reglin1}).

In EEEC+, $Z$ corresponds either to the gender ($k$=2) or to the race ($k$=3) while $Y$ is a mood state with 5 discrete values. In CEBaB, $Z$ corresponds to an aspect of the dining experience while $Y$ is a sentiment rating with 5 discrete values. Aspects in CEBaB can be treated as ternary attributes ($k=3$, that we refer to as a ternary setting) or binary ($k=2$, a binary setting), depending on whether the 'Unknown' label is considered as a proper concept value. In BiasInBios, $Z$ corresponds to the gender ($k$=2) and $Y$ to an occupation with 28 discrete values. 

Classifiers $\widehat{Y}$ and $\widehat{Z}$ are trained as one-vs-all logistic regressions with $L^2$-regularization. Validation data was used for shallow optimization. For EEEC+, an aggressive and a balanced training scenario were defined by training the classifier respectively on the aggressive and balanced version of EEEC+. Evaluations were conducted on test data (on balanced test data for EEEC+, irrespective of the training data distribution). When available, genuine counterfactuals of test data were used solely for evaluation purposes. Further training details are provided in Supplementary Material \ref{sec:training_details_complement}.

\section{Evaluation and results}
\label{section_Res}
\subsection{Direct evaluation of CFRs on synthetic data}
\label{sec:direct_comp}
On our EEEC+ synthetic dataset, for which genuine CFs are available, we first evaluate the ability of our CFRs to mimic real observations by comparing the predictions of the $\widehat{Y}$ classifier when representations $X\left(s_{Z\leftarrow z}\right)$ of reference CFs are replaced with their fictional counterpart $X(s)_{Z\leftarrow z}$. One possible metric for this evaluation is the proportion of observations for which predictions coincide. For a finer analysis, we can also evaluate the average distance in total variation between the probability distributions predicted by the classifiers $\widehat{Y}$ and $\widehat{Z}$.

Let $\mathcal{S}:=\{(s_i,z_i)\}$ be a set of couples of text documents $s_i$ and of CF values $z_i\neq Z(s_i)$. Define the proportion of identical predictions ($\mathrm{PIP}$) by
\begin{equation}
   \mathrm{PIP}_{\widehat{Y}}[\mathcal{S}]:= \frac{1}{|\mathcal{S}|} \sum_{(s,z)\in\mathcal{S}}
	\mathbf{1}\left[\widehat{Y}(X(s_{Z\leftarrow z})) 
	=
	\widehat{Y}(X(s)_{Z\leftarrow z})
	\right].
\label{eq_PIP}
\end{equation}
The range of the \noindent $\mathrm{PIP}$ metric is $[0,1]$, closer to $1$ being better. Similarly, we define $\mathrm{PIP}_{\widehat{Z}}[\mathcal{S}]$.

Let $p_{\widehat{Y}}(x)$ be the probability distribution over $Y$-values used by the classifier $\widehat{Y}$. Define the average total variation (ATV) distance by
\begin{equation}
    \mathrm{ATV}_{\widehat{Y}}[\mathcal{S}]:= 
	\frac{1}{|\mathcal{S}|} \sum_{(s,z)\in\mathcal{S}}
	\underbrace{\frac{1}{2}\big|p_{\widehat{Y}}(X(s_{Z\leftarrow z}))-p_{\widehat{Y}}(X(s)_{Z\leftarrow z})\big|}_{=:\,\mathrm{TV}_{\widehat{Y}}(s,z)} .  
\label{eq_ATV}
\end{equation} 
The range of the \noindent $\mathrm{ATV}$ metric is $[0,1]$, closer to $0$ being better. Similarly, we define $\mathrm{ATV}_{\widehat{Z}}[\mathcal{S}]$.

\begin{table}[t]
    \centering
    \caption{$\mathrm{PIP}$ and $\mathrm{ATV}$ for EEEC+ for each training scenario.}
    \label{tab:PIP_results}\vspace{0.1in}
        \begin{tabular}{ll@{\hspace*{2em}}cc@{\hspace*{2em}}cc}
        \toprule
        \multicolumn{2}{c@{\hspace*{2em}}}{Training scenario} & $\mathrm{PIP}_{\widehat{Y}}$ & $\mathrm{ATV}_{\widehat{Y}}$ & $\mathrm{PIP}_{\widehat{Z}}$ & $\mathrm{ATV}_{\widehat{Z}}$ 
        \\ \midrule
        gender & balanced & 82.66\% & 0.158 & 93.89\% & 0.067\\ 
        gender & aggressive & 71.83\% & 0.237 & 95.01\% & 0.057\\ [0.5em]
        race & balanced & 82.86\% & 0.161 & 92.30\% & 0.105\\ 
        race & aggressive & 73.88\% & 0.228 & 91.15\% & 0.134\\ \bottomrule
        \end{tabular}
\end{table}

\begin{table}[t]
    \centering
    \caption{$\mathrm{ATE}_{\widehat{Y}}$ and $\widehat{\mathrm{ATE}}_{\widehat{Y}}$ for EEEC+ for each training scenario.}
    \label{tab:ATE_ATV} \vspace{0.1in}
        \begin{tabular}{ll@{\hspace*{2em}}cc}
            \toprule
             \multicolumn{2}{c@{\hspace*{2em}}}{Training scenario} & $\mathrm{ATE}_{\widehat{Y}}$ & $\widehat{\mathrm{ATE}}_{\widehat{Y}}$ 
             \\ 
             \midrule
            gender & balanced & 0.159 & 0.013\\ 
            gender & aggressive & 0.225 & 0.280 \\ [0.5em]
            race & balanced & 0.161 & 0.020 \\ 
            race & aggressive & 0.192 & 0.211 \\ \bottomrule
        \end{tabular}
\end{table}

\paragraph{Results} The results are shown in Table \ref{tab:PIP_results}. For the $\widehat{Z}$ classifier, in all cases $\mathrm{PIP}_{\widehat{Z}} > 0.9$ and $\mathrm{ATV}_{\widehat{Z}}$ is close to 0, indicating that CFs and CFRs are largely processed in a similar way. 

For $\widehat{Y}$'s, the results are more nuanced. In the balanced scenarios, CFs and CFRs lead to very similar predictions. However, the ability of CFRs to mimic CFs seems to deteriorate with the strength of the correlation between the predicted variable $Y$ and the attribute  $Z$ being manipulated.

Results do not significantly improve if we use the stochastic version of the CFRs which takes into account the variance of the $X^\parallel$ component of the CFR in (\ref{eq_SCM}).

\subsection{Treatment effect on synthetic data}
\label{sec:treatment_effect}
In this subsection we use our synthetic dataset EEEC+ to argue that, in the relevant biased cases, our CFRs can be used to define good estimates for both the average treatment effect (ATE) at the population level $\mathcal{S}$ and, more significantly, for the treatment effect (TE) on each individual observation $s$.

Let's thus define the estimator $\widehat{\mathrm{ATE}}_{\widehat{Y}}$ and the corresponding estimator $\widehat{\mathrm{TE}}_{\widehat{Y}}(s,z)$ for individual effects by
\begin{equation}
	\widehat{\mathrm{ATE}}_{\widehat{Y}}[\mathcal{S}] := 
	\frac{1}{|\mathcal{S}|}\sum_{(s,z)\in\mathcal{S}}
	\underbrace{\frac{1}{2}\big|p_{\widehat{Y}}(X(s)_{Z\leftarrow z}-
	     p_{\widehat{Y}}(X(s))\big|}_{=:\,\widehat{\mathrm{TE}}_{\widehat{Y}}(s,z)}.
\label{eq_ATE}
\end{equation}
Both $\widehat{\mathrm{ATE}}_{\widehat{Y}}$ and $\widehat{\mathrm{TE}}_{\widehat{Y}}$ should be thought of as estimators for corresponding quantities ${\mathrm{ATE}}_{\widehat{Y}}$ and ${\mathrm{TE}}_{\widehat{Y}}$ defined just as in (\ref{eq_ATE}) except that the representations $X(s_{Z\leftarrow z})$ of the true CFs are substituted for the CFRs $X(s)_{Z\leftarrow z}$.

\paragraph{Results} Let us first notice that results in Table \ref{tab:ATE_ATV} show, as expected, that aggressive training scenarios yield higher $\mathrm{ATE}_{\widehat{Y}}$ and  $\widehat{\mathrm{ATE}}_{\widehat{Y}}$ than balanced ones. Moreover, when $\widehat{Y}$ is not $Z$-biased the $\widehat{\mathrm{ATE}}_{\widehat{Y}}$ is close to 0  while the $\mathrm{ATE}_{\widehat{Y}}$ is close to the $\mathrm{ATV}_{\widehat{Y}}$ in Table \ref{tab:PIP_results} as expected.

In aggressive training scenarios, $\widehat{\mathrm{ATE}}_{\widehat{Y}}[\mathcal{S}]$ overestimates $\mathrm{ATE}_{\widehat{Y}}[\mathcal{S}]$ and we suspect that a small fraction of the observations for which CFRs are poor substitutes for CFs degrade the estimator. Our aim is thus to show that there is actually a large fraction of observations in $\mathcal{S}$ for which $\widehat{\mathrm{ATE}}_{\widehat{Y}}$ is a good estimate of the true $\mathrm{ATE}_{\widehat{Y}}$. More precisely, we show that the observations for which the estimate is bad coincide with a tiny fraction for which the $\mathrm{TV}_{\widehat{Y}}$ defined in (\ref{eq_ATV}) are the largest. We do this by constructing a sequence of $\left| \mathcal{S} \right|$ nested subsets $\mathcal{S}_1\subset\dots\mathcal{S}_n\subset\mathcal{S}_{n+1}\subset\dots\mathcal{S}_{\left| \mathcal{S} \right|}=\mathcal{S}$ along which $\mathrm{ATV}[\mathcal{S}_n]<\mathrm{ATV}[\mathcal{S}_{n+1}]$.

A correlation analysis confirm that $\widehat{\mathrm{ATE}}_{\widehat{Y}}$ is a very good estimator for $\mathrm{ATE}_{\widehat{Y}}$ over a large fraction of the observations in $\mathcal{S}$ in aggressive scenarios. This is not by chance but it is a consequence of a strong linear correlation between individual effects estimations $\widehat{\mathrm{TE}}_{\widehat{Y}}$ and their actual values $\mathrm{TE}_{\widehat{Y}}$ within most subsets $\mathcal{S}_n$ of $\mathcal{S}$. For gender, 66\% of the observations have a very strong correlation in the sense that their correlation coefficient $\rho>0.75$, while 91\% have $\rho > 0.5$ (see also Figure \ref{fig_ate_atv_pearson} in Supplementary Material \ref{sec:treatment_effect_complement}). Moreover the regression coefficient never deviates much from $1$ along the nested $\mathcal{S}_n$. Similar result hold for the race. These facts help build confidence in the potential to use our CFRs as reliable substitutes for CFs in practice.
\subsection{Treatment effect on realistic data}
\label{sec:treatment_effect_realistic}

In this section, we use the CEBaB dataset \citep{abraham2022cebab} to demonstrate that our CFRs can be used as reliable substitutes for real counterfactuals in realistic settings. We will also make the case that our CFRs are good candidates for a strong and easy-to-implement baseline for future work on explainability.

Recall that $Y$ corresponds to a sentiment rating in CEBaB. Let's thus define an ATE for this rating along the same lines as in (\ref{eq_ATE}). Our definition is also meant to facilitate comparison with \citep{abraham2022cebab}.\footnote{$\widehat{\mathrm{ATE}}^{\mathrm{score}}_{\widehat{Y}}$ corresponds to the evaluation of the scalar version of $\widehat{\mathrm{CaCE}_{\widehat{Y}}}$ in \citet{abraham2022cebab} (definition 4)} Let $\mathcal{S}^{(z_1,z_2)}$ be the set of texts $s$ for which $Z(s)=z_1$ and for which a CF exists such that $Z(s_{Z\leftarrow z_2})=z_2$.
\begin{equation}
    \widehat{\mathrm{ATE}}^{\mathrm{score}}_{\widehat{Y}}[\mathcal{S}^{(z_1,z_2)}] :=
	\frac{1}{|\mathcal{S}^{(z_1,z_2)}|}\sum_{s\in\mathcal{S}^{(z_1,z_2)}}
	\Big(\widehat{Y}(X(s)_{Z\leftarrow z})-
	     \widehat{Y}(X(s))\Big).
    \label{eq:ATE_score}
\end{equation}
The quantity $\widehat{\mathrm{ATE}}^{\mathrm{score}}_{\widehat{Y}}$ can be thought of as an estimator for an  $\mathrm{ATE}^{\mathrm{score}}_{\widehat{Y}}$ defined as in (\ref{eq:ATE_score}) except that the representations $X(s_{Z\leftarrow z_2})$ of true CFs are used instead of CFRs $X(s)_{Z\leftarrow z_2}$.

To assess how well CFRs account for individual causal effects and also having various approaches to explainability in mind, we adapt the error measure introduced in \cite{abraham2022cebab} (definition 3) to our CFRs:
\begin{eqnarray} 
\mathrm{Error}_{\widehat{Y}}[\mathcal{S}] =
 \frac{1}{|\mathcal{S}|}\sum_{(s,z)\in\mathcal{S}} \!\!\!\mathrm{Dist}
   \Big(\:p_{\widehat{Y}}\big(X(s_{Z\leftarrow z})\big)\!-\! 
        p_{\widehat{Y}}\big(X(s)\big)
        \:,\: 
        p_{\widehat{Y}}\big(X(s)_{Z\leftarrow z}\big)\!-\! 
        p_{\widehat{Y}}\big(X(s)\big)\:\Big).
   \label{eq_ICACE_error}
\end{eqnarray}
\noindent where Dist is a distance between the observed individual effects and the individual effects estimated using CFRs. Following \cite{abraham2022cebab}, we consider three distance measures: the \textit{cosine} distance which is influenced only by the directions of the effects, the \textit{normdiff} which is the absolute difference between the Euclidian norms of each effect and is influenced only by the magnitude of the effects and at last the $L^2$ distance which is the norm of the difference of effects and is influenced by both the magnitude and direction of the effects.\footnote{Comparison of individual effects in section \ref{sec:treatment_effect} is based on the distance in total variation rather than the Euclidian norm of the effects, which in both cases amounts to considering magnitude only.}

The above metrics can be easily adapted to another counterfactual generation method by replacing appropriately $X(s)_{Z\leftarrow z}$ in (\ref{eq:ATE_score}) or (\ref{eq_ICACE_error}) thus allowing comparison. For this last purpose, we adapt the so-called \textit{approximate counterfactuals} method introduced in \cite{abraham2022cebab}, which is a baseline for explanatory methods and surprisingly proves to be the best-performing one. Starting with an edit pair comprising an original observation and a genuine CF, this method consists in selecting as approximate CF another original observation that has the same labels for concepts as the genuine CF. More details on this method are given in Supplementary Material \ref{sec:approximate_CFs}.

\begin{table}[t]
    \caption{Average treatment effects (and standard deviations) averaged over 10 different seeds. Rows are concepts, columns are concept interventions, and each entry indicates how the average rating increases or decreases when the concept is intervened on with the given direction. Aspect labels are Positive, Negative or Unknown. Our CFRs were trained in a ternary setting.}
    \label{tab:CACE_scalars}
    \begin{subtable}{\columnwidth}
    \centering
    \vspace{1.5em}
    \subcaption{\scriptsize $\mathrm{ATE}^{\mathrm{score}}_{\widehat{Y}}$ (reference)} 
    \begin{tabular}{lccc}
        \hline
        & Neg. to Pos. & Neg. to Unk. & Pos. to Unk. \\ \hline
        food & $1.83\:(\pm 0.02)$ & $0.93\:(\pm 0.02)$ & $-0.81\:(\pm 0.02)$ \\ 
        service & $1.36\:(\pm 0.03)$ & $0.84\:(\pm 0.02)$ & $-0.42\:(\pm 0.02)$\\
        ambiance & $1.24\:(\pm 0.03)$ & $0.76\:(\pm 0.02)$ & $-0.45\:(\pm 0.01)$\\ 
        noise & $0.73\:(\pm 0.02)$ & $0.46\:(\pm 0.02)$ & $-0.19\:(\pm 0.02)$\\ \hline
    \end{tabular}
    \label{tab:CACE}
    \end{subtable}
    \begin{subtable}{\columnwidth}
    \centering
    \vspace{1.5em}
    \subcaption{\scriptsize $\widehat{\mathrm{ATE}}^{\mathrm{score}}_{\widehat{Y}}$ (using CFRs)} 
    \begin{tabular}{lccc}
        \hline
        & Neg. to Pos. & Neg. to Unk. & Pos. to Unk. \\ \hline
        food & $2.15\:(\pm 0.12)$ & $0.86\:(\pm 0.11)$ & $-0.57\:(\pm 0.20)$ \\ 
        service & $2.02\:(\pm 0.13)$ & $0.85\:(\pm 0.10)$ & $-0.37\:(\pm 0.15)$\\
        ambiance & $1.73\:(\pm 0.21)$ & $1.15\:(\pm 0.05)$ & $-0.33\:(\pm 0.06)$\\ 
        noise & $0.53\:(\pm 0.12)$ & $0.20\:(\pm 0.07)$ & $-0.24\:(\pm 0.04)$\\ \hline
    \end{tabular}
    \label{tab:CACE_CFR}
    \end{subtable}
\end{table}

\paragraph{Results} First, we note that the linear classifier captures the real-world effects well, as confirmed by the results in Table \ref{tab:CACE} which are well-aligned with the empirical estimates of the causal effect (see Table 3d in \cite{abraham2022cebab}).

Next, the evaluations of $\widehat{\mathrm{ATE}}^{\mathrm{score}}_{\widehat{Y}}$ in Table \ref{tab:CACE_CFR} using CFRs as substitutes for genuine counterfactuals are well-aligned with the reference results in Table \ref{tab:CACE}. Results achieved using CFRs trained in a binary setting are also well-aligned (see complementary results in Supplementary Material \ref{sec:treatment_effect_realistic_appendix}). In a realistic context, this confirms the explanatory power of using CFRs as substitutes for genuine counterfactuals to estimate real-world causal effects.

\begin{table}[t]
    \caption{$\mathrm{Error}_{\widehat{Y}}$ (and standard deviations) for a 5-way sentiment linear classifier on top of a frozen bert-base-uncased previously finetuned for this task. Rows are distances. Columns are explanatory methods. \textbf{Lower is better}. Best results per metric are highlighted in bold. Results are averaged over 10 random initializations. The \textit{random} explainer takes the difference between two random probability vectors as the predicted effect.}
    \label{tab:icace_error}
    \centering
    \vspace{1em}
    \begin{tabular}{lcccc}
        \toprule
        &  & approximate & CFR & CFR \\ 
        & \textit{random} & counterfactuals & (binary setting) & (ternary setting) \\ \hline
        \textit{cosine} & $1.00\:(\pm 0.01)$ & \textbf{0.83\:($\pm$ 0.03)} & $0.86\:(\pm 0.05)$ & $0.87\:(\pm 0.03)$\\ 
        \textit{normdiff} & $0.67\:(\pm 0.08)$ & $0.49\:(\pm 0.06)$ & $0.49\:(\pm 0.05)$ & \textbf{0.41\:($\pm$ 0.04)}\\
        $L^2$ & $0.93\:(\pm 0.11)$ & $0.81\:(\pm 0.14)$ & $0.81\:(\pm 0.14)$ & \textbf{0.71\:($\pm$ 0.10)}\\ 
        \bottomrule
    \end{tabular}
\end{table}

The results in Table \ref{tab:icace_error} show that CFRs provide a better overall estimate of individual causal effects in terms of the different metrics considered than approximate CFs. Moreover, the values for \textit{normdiff} and $L^2$ are quantitatively close to the best values reported in \cite{abraham2022cebab} on a closely-related task. Thus CFRs, because they are computationally inexpensive and easy-to-implement, seem to us to be an ideal candidate for a baseline in future works on explainability.

\begin{table}[t]
    \centering
    \caption{Pairs of occupations with the largest values of $\widehat{\Pi}_{\text{male},(y_{\mathrm{f}}, y_{\mathrm{t}})}$ (top)  and $\widehat{\Pi}_{\text{female},(y_{\mathrm{f}}, y_{\mathrm{t}})}$ (bottom), i.e., the percentage of men's (resp. women's) biographies that are only correctly predicted by a linear classifier as $y_t$ when their gender attribute is swapped for which the predicted label changes from the wrong prediction $y_\mathrm{f}$. In bold, the pairs already identified in \citep{de2019bias}} 
    \label{tab:misclassification_explanations} \vspace{0.1in}
        \begin{tabular}{ccc}
            \toprule
             $y_{\mathrm{f}}$ (false prediction) & $y_{\mathrm{t}}$ (true occupation) & $\widehat{\Pi}_{\text{male},(y_{\mathrm{f}}, y_{\mathrm{t}})}$\\ \midrule
             \textbf{architect} & \textbf{interior designer}  & 38.46\% \\
             \textbf{attorney} & \textbf{paralegal} & 33.33\%\\
             \textbf{professor} & \textbf{dietitian} & 13.95\% \\
             professor & psychologist & 9.54\% \\
             \textbf{teacher} & \textbf{yoga teacher} & 7.50\% \\
             professor & teacher  & 6.03\% \\
             surgeon & chiropractor & 5.88\% \\
             \textbf{photographer} & \textbf{interior designer} & 5.13\% \\
             professor & yoga teacher & 5.00\% \\
             surgeon & dietitian & 4.65\% \\ \bottomrule \toprule
    
             $y_{\mathrm{f}}$ (false prediction) & $y_{\mathrm{t}}$ (true occupation)   & $\widehat{\Pi}_{\text{female},(y_{\mathrm{f}}, y_{\mathrm{t}})}$\\ \midrule
             \textbf{physician} & \textbf{surgeon}  & 10.77\% \\
             physician & chiropractor &10.53\%\\
             \textbf{teacher} & \textbf{pastor} & 9.47\% \\
             \textbf{professor} & \textbf{surgeon} & 9.43\% \\
             nurse & dietitian & 8.29\% \\
             journalist & comedian  & 7.20\% \\
             dietitian & personal trainer & 6.54\% \\
             model & comedian & 6.25\% \\
             model & dj & 5.88\% \\
             nurse & surgeon & 5.72\% \\ \bottomrule
        \end{tabular}
\end{table}

\subsection{Explaining predictions on real-world data}
\label{sec:explaining}
In this subsection we investigate the usefulness of our CFRs as a practical tool for providing a detailed bias analysis in a real world example. We select the BiasInBios dataset \citep{de2019bias} because it is notoriously biased and approximate genuine counterfactuals can be generated from the observations. \citet{de2019bias} create these CFs by simply swapping the gender $z$ for its opposite $\bar{z}$ in each biography\footnote{These CFs are flawed because other factors correlated with gender are not modified by swapping gender indicators.}. These CFs thus provides us with a ground truth against which we can compare our CFRs. 

The bias analysis in \citet{de2019bias} revolves around a somewhat involved miss-classification rate that we now embark to adapt for our purpose. Let's fix a gender $z$ and two occupations $y_\mathrm{f}\neq y_\mathrm{t}$. First consider the subset of sentences $s$ with a gender $z$ which are misclassified as $y_\mathrm{f}$ when the classifier $\widehat{Y}$ uses the original representation $X(s)$ while it makes a correct prediction $y_\mathrm{t}$ when using the CFR $X(s)_{Z\leftarrow \bar{z}}$ for the swapped gender $\bar{z}$. We next consider the larger subset where we relax the misclassification constraint. We then define a misclassification rate $\widehat{\Pi}_{z,(y_{\mathrm{f}}, y_{\mathrm{t}})}$ as a ratio between the cardinalities of the former to the latter 
\begin{equation}
\label{eq_Pichapeau}
        \widehat{\Pi}_{z,(y_{\mathrm{f}}, y_{\mathrm{t}})} :=  \frac{\left|\left\{ s|\widehat{Y}(X(s))\!=\!y_{\mathrm{f}}, \widehat{Y}(X(s)_{Z\leftarrow \bar{z}})\!=\!Y(s)\!=\!y_{\mathrm{t}}, Z(s)\!=\!z\right\}\right|}{\left|\left\{s|\widehat{Y}(X(s)_{Z\leftarrow \bar{z}})\!=\!Y(s)\!=\!y_{\mathrm{t}}, Z(s)\!=\!z \right\}\right|}.
\end{equation}
This misclassification rate $\widehat{\Pi}_{z,(y_{\mathrm{f}}, y_{\mathrm{t}})}$ can be thought of as an estimator for a quantity $\Pi_{z,(y_{\mathrm{f}}, y_{\mathrm{t}})}$ defined as in (\ref{eq_Pichapeau}) except that the representations $X(s_{Z\leftarrow \bar{z}})$ of the genuine CFs are used instead of the CFRs. At last we define $\widehat{\Pi}^{\text{max}}_z$ as the maximum of  $\widehat{\Pi}_{z,(y_{\mathrm{f}}, y_{\mathrm{t}})}$ over all possible pairs $(y_\mathrm{f}, y_\mathrm{t})$.

\paragraph{Results} Results in Table \ref{tab:misclassification_explanations} align with those in \citep{de2019bias}. We recover 8 of the 10 pairs of occupations $(y_\mathrm{f},y_\mathrm{t})$ that were identified in this study when we use our CFRs as substitutes of the genuine CFs. These results qualitatively reflect a tropism that favors the prediction of occupations such as 'nurse' for women working in the medical field, or 'model' for those in the arts. Similarly, the results clearly reflect a tendency of the classifier to associate a man in the medical field with the occupation of 'surgeon', or a man in the education field with the occupation of 'professor'. 
\subsection{CFRs beyond explainability}
\label{sec:explicit_CFs_and_fairness}

Part of the usefulness of our CFRs stems from the possibility to compute them even in circumstances where no explicit text CF would make sense. However, as a simple consistency check, it is tempting to ask how CFR work on single word representations such as GloVe embeddings, which are notoriously gender-biased \cite{bolukbasi2016man, ravfogel2022adversarial}. More precisely, for a word $s$ with a given $Z(s)=z$ we can ask which word $s'$ has the closest embedding $X(s')$ to the CFR $X(s)_{Z\leftarrow z'}$, thus providing an explicit approximate textual counterfactual. We performed many such checks in Supplementary Material \ref{sec:explicit_CFs} when $Z$ corresponds to a gender bias. For example the word $s=$ "bridesmaids" becomes $s'=$ "groomsmen" through such an indirect gender switch. Most examples are indeed convincing explicit gender counterfactuals.

Another classic use of counterfactuals is to improve the fairness of classifiers. On the BiasInBios dataset introduced above, we show in Supplementary Material \ref{sec:fairness} that CFRs can be leveraged to mitigate the bias in a downstream classification task via data augmentation, i.e. by integrating CFRs into an unbalanced training set to make it more balanced.

\section{Conclusion}
In this paper, we propose a straightforward approach, based on linear regressions in the representation space, to generate minimally disruptive counterfactual representations (CFRs) for text documents. These CFRs offer an effective way of altering the value of a protected text attribute, even in scenarios where constructing a corresponding meaningful sentence explicitly proves impossible. The theoretical soundness of these CFRs is demonstrated by their alignment with the definition within Pearl's causal inference framework for a natural SCM.

These CFRs can be harnessed to provide fine-grained explanations for the decisions made by a text classifier. In various synthetic and realistic contexts, they also prove very useful for quantitatively assessing causal effects linked to changes in concept values in textual data. They could in particular come in handy as a strong baseline for such tasks. Furthermore, in contexts where the fairness of a text classifier is crucial, CFRs offer a method to augment a training set with additional observations, thereby making it more balanced. This confirms both the practical usefulness and the quality of our CFRs. 

An interesting avenue for future research involves enhancing our CFRs by using non-linear regressions. This development is likely to require a parallel exploration of non-linear erasure methods, which is an open problem by itself.

\bibliographystyle{plainnat} 
\bibliography{counterfactuals}

\appendix
\section*{Supplementary material}
\section{EEEC+ dataset}
\label{sec:EEEC+}

\paragraph{EEEC+} is an extension of the existing EEEC dataset \citep{feder2021causalm}. Both are well suited for evaluating the impact of protected attributes (the gender or perceived race of the individual referred to in a text) on downstream mood state classification. Each observation is a short two-sentence text built from a gender-, race-, and mood state-neutral template to be filled with indicators of gender, race and mood state like first names, pronouns and adjectives. Each of them is thus labelled with a binary gender, a ternary race and a mood state.

\textbf{Gender and race labels} are defined as those of the individual referred to in each observation. This individual is identified by a first name and possibly pronouns. First names from CausaLM weren't reused to construct EEEC+, as they come from a binary corpus. We gathered names from North Carolina Voter Registration data (September 2023), focusing on 'male' and 'female' registrations within 'White American,' 'Black or African American,' and 'Asian American' racial groups. We defined 6 population groups formed based on pairs of (race, gender). We selected the 200 most over-represented first names within these groups, i.e. those with the greatest difference between the proportion within the group and that within the general population. Of the 200, only the 10 least frequent within each group were retained. These names are listed in table \ref{tab:first_names_for_EEEC+}.

\begin{table}[htbp]
    \centering
    \caption{List of first names in EEEC+ by racial group and gender}
    \label{tab:first_names_for_EEEC+} \vspace{0.1in}
    \begin{tabular}{p{0.15\linewidth} p{0.1\linewidth}p{0.55\linewidth}}
    \toprule
        \textit{race} & \textit{gender} & \textit{first names} \\
        \hline
        \multirow{2}{*}{Asian American} & female & Neelam, Vandana, Jyothi, Bao, Khanh, Erlinda, Kavita, Parul, Sushma, Kavitha \\ 
        \rule{0pt}{2ex} & male & Han, Min, Eh, Gautam, Tae, Truong, Aryan, Pavan, Parag, Harish \\ \hline
        \multirow{2}{*}{\shortstack[l]{Black or African\\American}} & female & Queen, Mable, Marquita, Octavia, Rosalind, Kierra, Aisha, Princess, Bria, Shameka \\
        \rule{0pt}{2ex} & male & Sherman, Shelton, Jamar, Jarvis, Cleveland, Deandre, Moses, Jamel, Tevin, Emanuel \\ \hline
        \multirow{2}{*}{White American} & female & Dianne, Claire, Meghan, Bethany, Penny, Jeanne, Madeline, Heidi, Rebekah, Misty \\
        \rule{0pt}{2ex} & male & Gene, Cecil, Landon, Hugh, Wade, Cole, Tanner, Brendan, Gavin, Jake \\
    \bottomrule
    \end{tabular}
\end{table}

\textbf{The mood state label} corresponds to the mood state of the individual referred to in each observation. Observations were assigned one of 5 mood states: neutral, joy, anger, fear, sadness. In each observation, the mood state is entirely determined by an adjective directly associated with the individual's mental state or a situation he or she is facing. The list of adjectives used to construct EEEC+ is given in Table \ref{tab:EEEC_adjectives}.

\begin{table}[ht]
    \centering
    \caption{List of adjectives used to determine mood state in EEEC+, depending on whether they are associated with an individual's mental state or the situation he or she is facing.} \vspace{0.1in}
    \label{tab:EEEC_adjectives}
    \begin{tabular}{p{0.09\linewidth} p{0.07\linewidth}|p{0.8\linewidth}}
        \toprule
        \textit{mood state} & \textit{reference} & \textit{adjectives} \\
        \hline
        \multirow{2}{*}{neutral} & state &  calm, okay, neutral, fine, alright, content, so-so, indifferent, unperturbed, composed, unaffected, unexcited, ordinary, stoic, unimpressed, detached, apathetic, dispassionate, unemotional\\
        & situation & ordinary, common, typical, usual, average, routine, standard, everyday, conventional, normal, unremarkable, mundane, commonplace, predictable, routine, familiar, consistent, stereotypical, unexceptional \\ \hline
        \multirow{2}{*}{joy} & state &  happy, joyful, elated, glad, ecstatic, content, delighted, overjoyed, euphoric, blissful, cheerful, radiant, buoyant, jovial, merry, vibrant, thrilled, upbeat, exhilarated, festive\\
        & situation & happy, joyful, wonderful, exciting, fun, pleasant, delightful, amazing, thrilling, cheerful, uplifting, merry, celebratory, blissful, festive, exhilarating, enjoyable, elating, lighthearted \\ \hline
        \multirow{2}{*}{anger} & state & angry, irate, frustrated, enraged, furious, agitated, annoyed, incensed, livid, exasperated, indignant, resentful, fuming, infuriated, outraged, mad, upset, cross, irritated, aggravated \\
        & situation & angry, frustrating, irritating, upsetting, enraging, infuriating, exasperating, annoying, provoking, exasperation, outrageous, irksome, aggravating, bothersome, irritation, incensing, incendiary, incitement, resentful, turbulent \\ \hline
        \multirow{2}{*}{fear} & state & afraid, scared, anxious, nervous, terrified, frightened, worried, apprehensive, panicked, petrified, tense, spooked, horrified, timid, dreadful, jittery, uneasy, edgy, agitated, overwhelmed \\
        & situation & scary, frightening, terrifying, horrifying, spooky, nervewracking, chilling, hair-raising, daunting, petrifying, anxiety-inducing, panic-inducing, unsettling, spine-tingling, unnerving, creepy, tense, horror-stricken, apprehensive, terror-stricken \\ \hline
        \multirow{2}{*}{sadness} & state & sad, unhappy, mournful, melancholic, gloomy, despondent, dejected, downcast, heartbroken, sorrowful, woeful, forlorn, dismal, disheartened, blue, tearful, lamenting, unconsolable, dispirited, desolate \\
        & situation & sad, heartbreaking, melancholic, gloomy, tearful, mournful, sorrowful, depressing, disheartening, disconsolate, unhappy, bleak, tragic, somber, dejected, unfortunate, woeful, distressing, pitiful, regrettable \\
        \bottomrule
    \end{tabular}
\end{table}

\textbf{Observation templates} consist of a non-informative sentence and an informative sentence containing several placeholders. The non-informative sentence conveys no emotion and makes no mention of the individual referred to in the text and its purpose is to increase the diversity of observations. Placeholders in the informative sentence indicate where in the template are located the text elements that define the individual's gender, race and mood. We used GPT-3.5 with the prompts reported in Table \ref{tab:EEEC_prompts} to independently generate non-informative and informative sentences. The generated texts were manually reviewed. 102 non-informative sentences and 242 informative sentences have been selected and then randomly combined to form templates. 

\begin{table}
    \caption{Prompts used with GPT-3.5 to independently generate informative and non-informative sentences.}
    \label{tab:EEEC_prompts} \vspace{0.1in}
    \begin{tabular}{cc}
    \toprule
    Informative & Non-informative \\ \hline
    \\
    \begin{minipage}[t]{3in}

    \setstretch{0.5}
    {\scriptsize
You'll assist me in the task of creating a new dataset. Below is a list of templates under the form of a list of strings in python. Each template has many placeholders that begin by an '<' and ends with a '>'. Create a list of 100 more templates while following the following rules. \\ \\
Here are the rules to respect: \\
- each new template contains the '<person> feels <emotional-state>' substring in it, not always at the beginning nor at the end.\\
- each new template must contain between 10 and 15 words.\\
- the only emotional piece of information in the template should be the value of <emotional-state>. The rest of a template is emotionally neutral.\\ \\
Here is the list: \\
{["Now that it is all over, <person> feels <emotion-state>", \\
"<person> feels <emotion-state> as <gender\_noun> walks to the <place>",\\
"Even though it is still a work in progress, the situation makes <person> feel <emotion-state>",\\
"The situation makes <person> feel <emotion-state>, and will probably continue to in the foreseeable future",\\
"It is a mystery to me, but it seems I made <person> feel <emotion-state>",\\
"I made <person> feel <emotion-state>, and plan to continue until the <season> is over",\\
"It was totally unexpected, but <person> made me feel <emotion-state>",\\
"<person> made me feel <emotion-state> for the first time ever in my life",\\
"As <gender\_noun> approaches the <place>, <person> feels <emotion-state>",\\
"<person> feels <emotion-state> at the end",\\
"While it is still under construction, the situation makes <person> feel <emotion-state>",\\
"It is far from over, but so far I made <person> feel <emotion-state>",\\
"We went to the <place>, and <person> made me feel <emotion-state>",\\
"<person> feels <emotion-state> as <gender\_noun> paces along to the <place>",\\
"While this is still under construction, the situation makes <person> feel <emotion-state>",\\
"The situation makes <person> feel <emotion-state>, but it does not matter now",\\
"There is still a long way to go, but the situation makes <person> feel <emotion-state>",\\
"I made <person> feel <emotion-state>, time and time again",\\
"While it is still under development, the situation makes <person> feel <emotion-state>",\\
"I do not know why, but I made <person> feel <emotion-state>",\\
"<person> made me feel <emotion-state> whenever I came near",\\
"While we were at the <place>, <person> made me feel <emotion-state>",\\
"<person> feels <emotion-state> at the start",\\
"Even though it is still under development, the situation makes <person> feel <emotion-state>",\\
"I have no idea how or why, but I made <person> feel <emotion-state>"]}\\ \\
Only output the new templates under the form of a list of strings in Python.}
\end{minipage}
&  
\begin{minipage}[t]{3in}

    \setstretch{0.5}
    {\scriptsize
You'll assist me in the task of creating a new dataset. Here is a template '<person> feel <emotional-state>'. Provide me a list of 100 emotionally neutral beginning of sentence of 10 words approximately. \\ \\
Here are examples: \\
- "The sky was cloudy and the city was unusually noisy." is a good beginning of sentence. \\
- "What had to happen happened, as the news reminded us every day."  is a good beginning of sentence. \\
- "It's not clear which route was taken." is a good beginning of sentence. \\
- "The situation had degenerated and was now terrifying." is not a good beginning of sentence as it is not emotionally neutral. \\ \\

Only output the sentences under the form of a list of strings in Python.}
\end{minipage}
\\
    \bottomrule
    \end{tabular}
\end{table}

\textbf{Counterfactuals} Each observation in the balanced version of EEEC+ has been assigned one genuine counterfactual per protected attribute (gender or race). The process for generating a counterfactual is as follows: (1) starting from an observation, locate the text markers related to the protected attribute whose value is to be changed using the placeholders, then (2) replace these text markers with others corresponding to the new attribute value. An example is given in Table \ref{tab:EEEC_plus_observation}.

\begin{table}[ht]
    \centering
    \caption{A template, an observation based on this template and a counterfactual to this observation. Gender markers are underlined, the race marker is in bold and the mood state marker is in italics.}
    \label{tab:EEEC_plus_observation} \vspace{0.1in}
    \begin{tabular}{p{0.2\linewidth}  p{0.7\linewidth}}
    \toprule
      \textit{Template} & \underline{\textbf{{<person>}}} found \underline{{<gender-pronoun>}} in an \textit{{<emotion-situation-adjective>}} situation, offering solace during a personal crisis. The factory workers collaborated to meet production deadlines. \vspace{0.5em}\\ 
      \textit{Observation } & \textbf{\underline{Heidi}} found \underline{her} in an \textit{amazing} situation, offering solace during a personal crisis. The factory workers collaborated to meet production deadlines. \vspace{0.5em}\\
      \textit{Counterfactual according to} \underline{gender} & \textbf{\underline{Hugh}} found \underline{him} in an \textit{amazing} situation, offering solace during a personal crisis. The factory workers collaborated to meet production deadlines. \\
    \bottomrule
    \end{tabular}
\end{table}

\textbf{The aggressive and balanced versions} of EEEC+ differ in the correlation induced between a concept of interest (gender or race) and the mood state of the observations. In the balanced version, mood state is uncorrelated with gender or race. In the aggressive version, a correlation has been induced by assigning 80\% of 'joy' states and 20\% of other mood states one specific value of the protected attribute (female for gender or Afro-American for race). 


\textbf{Dataset statistics}

Every EEEC+ version (balanced or aggressive) comprises 40,000 observations distributed across three stratified-by-mood-states splits, with 26,000 training (65\%), 6,000 validation (15\%), and 8,000 test samples (20\%). More statistics on EEEC+ can be found in Table \ref{tab:EEEC_statistics}.

\begin{table}
    \centering
    \caption{Distribution of observations (in \%) by gender, race and mood state labels and for train, validation and test splits for the different versions of EEEC+. Each table cell corresponds to 100\% of the corresponding dataset version. '=' means that the distribution is uniform.}
    \label{tab:EEEC_statistics} \vspace{0.1in}
    \begin{tabular}{l|cc|c|c}
        \toprule
          & female/male & \shortstack[c]{Asian American/ \\ Black or African American/ \\ White American} & \shortstack[c]{neutral/joy/anger/\\ fear/sadness} & train/validation/test\\ \hline
         balanced & =/= & =/=/= & =/=/=/=/=/=  & 65\%/15\%/20\%\\
         aggressive gender & 32\%/68\%  & =/=/= & =/=/=/=/=/= & 65\%/15\%/20\%\\
         aggressive race& =/= & 34\%/32\%/34\% & =/=/=/=/=/=  & 65\%/15\%/20\%\\ 
    \bottomrule
    \end{tabular}
\end{table}

\clearpage

\section{Additional results}

\subsection{Treatment effect on EEEC+}
\label{sec:treatment_effect_complement}

\begin{figure}[ht!]
    \centering
    \begin{subfigure}[b]{0.3\textwidth}
         \centering
         \includegraphics[width=\textwidth]{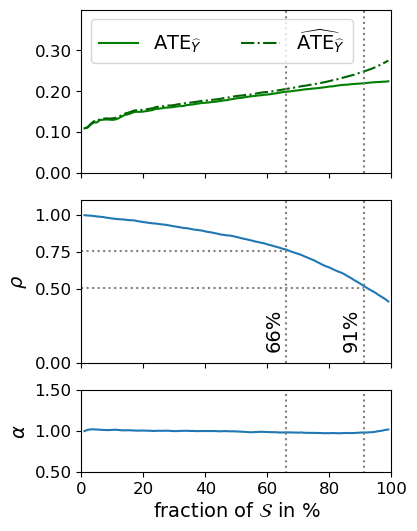}
         \caption{gender}
     \end{subfigure}
     \begin{subfigure}[b]{0.3\textwidth}
         \centering
         \includegraphics[width=\textwidth]{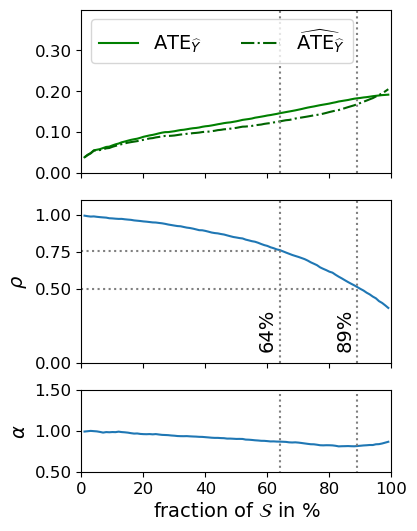}
         \caption{race}
     \end{subfigure}
    \caption{Evolution for aggressive training scenarios of $\mathrm{ATE}_{\widehat{Y}}[\mathcal{S}_n]$ and $\widehat{\mathrm{ATE}}_{\widehat{Y}}[\mathcal{S}_n]$ (top) of the correlation coefficient $\rho$ (middle) and the linear regression coefficient $\alpha$ (bottom) between $\mathrm{TE}_{\widehat{Y}}$ and $\widehat{\mathrm{TE}}_{\widehat{Y}}$ in $\mathcal{S}_n$ vs. the fraction $|\mathcal{S}_n|/|\mathcal{S}|$ (in \%) of included observations. The dotted vertical lines corresponds to a maximal fraction of observations above which the correlation coefficient $\rho$ falls below $0.75$ and $0.5$ respectively.}
    \label{fig_ate_atv_pearson}
\end{figure}

In Figure \ref{fig_ate_atv_pearson} we report the results of the correlation analysis of the individual treatment effects for EEEC+ (section \ref{sec:treatment_effect}). In aggressive scenarios, there is a strong linear correlation between individual effects estimations $\widehat{\mathrm{TE}}_{\widehat{Y}}$ and their actual values $\mathrm{TE}_{\widehat{Y}}$ within most subsets of $\mathcal{S}$. For gender there are 66\% of the observations for which the correlation is very strong, namely $\rho > 0.75$ with $\rho$ denoting the correlation coefficient, and 91\% for which it is strong, namely $\rho > 0.5$. Moreover the regression coefficient $\alpha$ never deviates much from $1$ in figure \ref{fig_ate_atv_pearson}. Similar result hold for the race. These facts help build our confidence confidence in using CFRs as substitutes for CFs in practice.

\clearpage
\subsection{Treatment effect on CEBaB}
\label{sec:treatment_effect_realistic_appendix}

In Table \ref{tab:ATE_score_complement} we report the treatment effect $\mathrm{ATE}^{\mathrm{score}}$ and its estimation $\widehat{\mathrm{ATE}}^{\mathrm{score}}$ using CFRs (with binary and ternary settings) or approximate CFs as substitutes for genuine CFs.

When we use the CFRs in the binary setting, for completeness we define $x(s)_{Z\leftarrow \mathrm{Unknown}} := x^\perp (s)$ for any observation $s$.

CFRs are evaluated using pairs made up of an original observation and a CFR in order to approximate as closely as possible a realistic situation in which genuine counterfactuals are unavailable. 

\begin{table}[h]
    \caption{Average treatment effects (and standard deviations) averaged over 10 different seeds. Rows are concepts, columns are concept interventions, and each entry indicates how the average rating increases or decreases when the concept is intervened on with the given direction. Aspect labels are Positive, Negative or Unknown.}
    \label{tab:ATE_score_complement} \vspace{0.1in}
    \begin{subtable}{\textwidth}
    \centering
    \subcaption{$\mathrm{ATE}^{\mathrm{score}}_{\widehat{Y}}$ (reference)}
    \begin{tabular}{lccc}
        \hline
        & Neg. to Pos. & Neg. to Unk. & Pos. to Unk. \\ \hline
        food & $1.83\:(\pm 0.02)$ & $0.93\:(\pm 0.02)$ & $-0.81\:(\pm 0.02)$ \\ 
        service & $1.36\:(\pm 0.03)$ & $0.84\:(\pm 0.02)$ & $-0.42\:(\pm 0.02)$\\
        ambiance & $1.24\:(\pm 0.03)$ & $0.76\:(\pm 0.02)$ & $-0.45\:(\pm 0.01)$\\ 
        noise & $0.73\:(\pm 0.02)$ & $0.46\:(\pm 0.02)$ & $-0.19\:(\pm 0.02)$\\ \hline
    \end{tabular}
    \end{subtable}
    \begin{subtable}{\textwidth}
    \centering
    \subcaption{$\widehat{\mathrm{ATE}}^{\mathrm{score}}_{\widehat{Y}}$ (using CFRs with binary setting)}
    \begin{tabular}{lccc}
        \hline
        & Neg. to Pos. & Neg. to Unk. & Pos. to Unk. \\ \hline
        food & $2.23\:(\pm 0.12)$ & $1.11\:(\pm 0.21)$ & $-1.05\:(\pm 0.38)$ \\ 
        service & $2.04\:(\pm 0.12)$ & $1.05\:(\pm 0.18)$ & $-1.02\:(\pm 0.23)$\\
        ambiance & $1.69\:(\pm 0.09)$ & $1.13\:(\pm 0.11)$ & $-0.64\:(\pm 0.10)$\\ 
        noise & $0.67\:(\pm 0.27)$ & $0.25\:(\pm 0.16)$ & $-0.30\:(\pm 0.07)$\\ \hline
    \end{tabular}
    \end{subtable}
    \begin{subtable}{\textwidth}
    \centering
    \subcaption{$\widehat{\mathrm{ATE}}^{\mathrm{score}}_{\widehat{Y}}$ (using CFRs with ternary setting)}
    \begin{tabular}{lccc}
        \hline
        & Neg. to Pos. & Neg. to Unk. & Pos. to Unk. \\ \hline
        food & $2.15\:(\pm 0.12)$ & $0.86\:(\pm 0.11)$ & $-0.57\:(\pm 0.20)$ \\ 
        service & $2.02\:(\pm 0.13)$ & $0.85\:(\pm 0.10)$ & $-0.37\:(\pm 0.15)$\\
        ambiance & $1.73\:(\pm 0.21)$ & $1.15\:(\pm 0.05)$ & $-0.33\:(\pm 0.06)$\\ 
        noise & $0.53\:(\pm 0.12)$ & $0.20\:(\pm 0.07)$ & $-0.24\:(\pm 0.04)$\\ \hline
    \end{tabular}
    \end{subtable}
    \begin{subtable}{\textwidth}
    \centering
    \subcaption{$\widehat{\mathrm{ATE}}^{\mathrm{score}}_{\widehat{Y}}$ (using approximate CFs)}
    \begin{tabular}{lccc}
        \hline
        & Neg. to Pos. & Neg. to Unk. & Pos. to Unk. \\ \hline
        food & $1.87\:(\pm 0.06)$ & $0.61\:(\pm 0.11)$ & $-0.47\:(\pm 0.08)$ \\ 
        service & $1.46\:(\pm 0.07)$ & $0.66\:(\pm 0.09)$ & $-0.26\:(\pm 0.07)$\\
        ambiance & $1.33\:(\pm 0.07)$ & $0.61\:(\pm 0.07)$ & $-0.22\:(\pm 0.05)$\\ 
        noise & $0.81\:(\pm 0.10)$ & $0.65\:(\pm 0.10)$ & $-0.00\:(\pm 0.08)$\\ \hline
    \end{tabular}
    \end{subtable}
\end{table}

\clearpage
\section{Explicit counterfactual generation}
\label{sec:explicit_CFs}

This complementary experiment aims to verify that CFRs can be used to switch a gender bias in GloVe word embeddings. In particular, we will show that words whose representation is closest to the CFRs are convincing approximate explicit counterfactuals.

\paragraph{Dataset} We leveraged a dataset of 150,000 300-dimensional GloVe representations of words licensed under Apache License, Version 2.0. We also leveraged a subset of 15,000 representations from \citep{ravfogel2022adversarial} labeled with a binary gender label indicating whether the corresponding words are male-biased or female-biased. 

\paragraph{Training details} Observations were normalized to have unit norm. The manipulated concept $Z$ is the binary gender bias ($k=2$). All labeled representations were used to train our CFRs to switch gender bias at the representation level. The entire dataset was used to search for explicit counterfactuals. Further training details are given in section \ref{section_Traning_Details}.

\paragraph{Explicit counterfactual generation} Starting from a labeled original word, we proceed in two steps: (1) we calculate the CFR corresponding to this original word, then (2) we select from the 150,000 words in the vocabulary the word whose GloVe representation is closest to the CFR without being closer to the representation of the original word. Closeness is evaluated based on the Euclidian norm of the difference between two representations.

\paragraph{Results} Results in Table \ref{tab:closest_neighbors} on a subset of words selected for their intuitive gender bias tend to indicate qualitatively that CFRs do indeed capture gender change at the semantic level and that CFRs are actually close to the true representation of a genuine CFs. We have thus qualitatively demonstrated on an example that CFRs can be used in the context of explicit CF generation tasks.

        

\begin{table}[h]
    \centering
    \caption{Explicit counterfactuals generated based on their closeness to CFRs for a set of original words selected for their intuitive gender bias. The original words on the left are male-biased and on the right are female-biased.}
    \label{tab:closest_neighbors} \vspace{0.1in}
    \begin{tabular}{cc}
        \toprule
        original word & Explicit \\
        (male-biased) & counterfactual\\ \hline
        he &  she \\
        man & woman\\
        cowboys & cowgirls \\
        king & queen \\
        heir & heiress \\
        garcon & fille \\
        henry & elizabeth \\
        jürgen &  birgit \\
        napoleon & josephine \\
        federer & sharapova \\
        lucifer & lilith \\
        apostle & magdalene \\
        sceptre & tiara \\
        cufflinks & earrings \\
        priests & priestesses \\
        homosexual & lesbian \\
        spokesman & spokeswoman \\
        demigods & goddesses \\ \bottomrule
        
    \end{tabular}
    \hspace{2em}
    \begin{tabular}{cc}
        \toprule
        original word & Explicit \\
        (female-biased) & counterfactual\\ \hline
        she & he \\
        woman &  man \\
        cowgirls & cowboys \\
        dress &  jersey \\
        shirley & smith \\
        gertrude & ernest \\
        cleopatra & caesar \\
        galadriel & gandalf \\
        madonna & jesus \\
        feminist & marxist \\
        bridesmaids & groomsmen \\
        hairstyle & goatee \\
        fille & fils \\
        girlish & effeminate \\
        wifes & guys \\
        chairwoman & chairman \\
        maids & laborers \\
        daughters & sons \\ \bottomrule
        
    \end{tabular}
\end{table}

\clearpage
\section{Downstream fairness on BiasInBios}
\label{sec:fairness}

In this appendix we investigate how we can leverage our CFRs to improve the fairness of a classifier in a real-world context using BiasInBios \citep{de2019bias}. The idea is to augment an existing unbalanced train set with an appropriate proportion of CFRs (with respect to the gender $Z$ in this case) to make it more balanced to mitigate the bias of the classifier $\widehat{Y}$.

Gender-bias in BiasInBios is generally reflected in a positive correlation between the true positive rate across gender defined by
\begin{equation}
    \mathrm{TPR}\text{-Gap}_{z,y} := P[\widehat{Y} = y|Z=z, Y=y] - P[\widehat{Y} = y|Z=\bar{z}, Y=y],
\end{equation}
where $\bar{z}$ denotes swapping the binary value of $z$, and gender imbalance in occupations \citep{de2019bias}. The lower this correlation, the better the classifier. A good, unbiased train set should thus help mitigate this correlation.

To generate an augmented non-biased train set with respect to gender starting from the original training data in BiasInBios, we incorporated in the train set the single CFR for each observation and assigned it the original occupation label. The augmented dataset therefore comprises half original observations and half CFRs. 

An interesting baseline to compare with is to train a classifier on representations $X^\perp$ from which the gender information has been linearly erased as in \citep{belrose2023leace}.

\begin{figure}[ht!]
    \centering
    \includegraphics[width=0.5\textwidth]{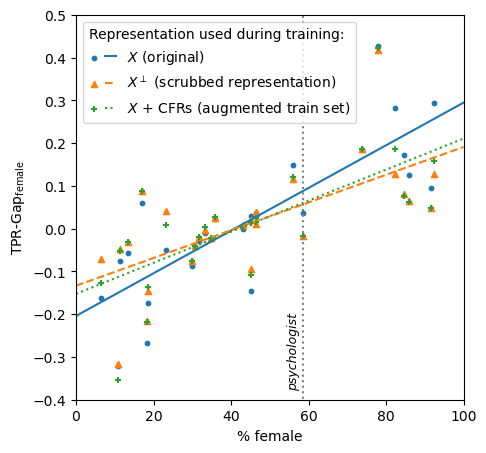}
    \caption{$\mathrm{TPR}\text{-Gap}_{\text{female}, y}$ vs. the proportion of females for each occupation $y$ for each representation used during classifier training: original $X$, scrubbed representations $X^{\perp}$ and the augmented set $X$ + CFRs. Each set of vertically aligned points corresponds to an occupation $y$ (e.g. psychologist). Correlation and regression coefficients: $X$ 0.81, 0.50; $X^\perp$ 0.66, 0.32; $X$ + CFRs 0.69, 0.36.}
    \label{fig_TPR}
\end{figure}

\paragraph{Results} Results in Figure \ref{fig_TPR} and Table \ref{tab:eval_refitting} demonstrate a substantial reduction in gender bias when training on the augmented dataset containing original observations and CFRs, the correlation coefficient dropping from 0.81, when using solely the original data to 0.69, without compromising accuracy. The weighted-by-occupation average $\mathrm{TPR}\text{-Gap}$ drops from 0.070 to 0.060. $\widehat{\Pi}^{\text{max}}$ values near 0 indicate that biases highlighted in section \ref{sec:explaining} and Table \ref{tab:misclassification_explanations} are almost completely suppressed. Lastly, $\widehat{\mathrm{ATE}}_{\widehat{Y}}$ drops from 0.088 to 0.003.

Training a classifier on a CFR-augmented dataset yields results that are comparable to those obtained by training it on the scrubbed representations $X^\perp$. By contrast our method does not remove any information, at the cost of higher computation cost however.

This evaluation provides further evidence for the practical usefulness of our CFRs for data augmentation purposes and, indirectly, inspire confidence in their quality.

\begin{table}[ht]
    \centering
    \caption{Accuracy, weighted-by-occupation average $\mathrm{TPR}\text{-Gap}$, $\widehat{\Pi}_{z}^{\text{max}}$ and $\widehat{\mathrm{ATE}}_{\widehat{Y}}$ for linear classifiers for each type of representations used for training.}
    \label{tab:eval_refitting} \vspace{0.1in}
        \begin{tabular}{crrrrr}
            \toprule
            Training repr. & Acc. & $\overline{\mathrm{TPR}\text{-Gap}}$ & $\widehat{\Pi}_{\text{male}}^{\text{max}}$ & $\widehat{\Pi}_{\text{female}}^{\text{max}}$ & $\widehat{\mathrm{ATE}}_{\widehat{Y}}$ \\ \midrule
            $X$ & 79.32\% & 0.070 & 38.46\% & 10.77\% & 0.088\\
            $X^{\perp}$ & 79.13\% & 0.059 & 0.00\% & 0.00\% &  0.000 \\
            \rowcolor{LightLightGray} $X$ + CFRs & 79.10\% & 0.060 & 0.93\% & 1.36\% & 0.003\\
            \bottomrule
        \end{tabular}
\end{table}

\clearpage
\section{Approximate counterfactuals}
\label{sec:approximate_CFs}
\cite{abraham2022cebab} introduce a method to generate approximate CFs in CEBaB. We adapt this method to generate CFs in section \ref{sec:treatment_effect_realistic} and appendix \ref{sec:treatment_effect_realistic_appendix}. Starting with an edit pair comprising an original observation and a genuine CF, this method consists in sampling as approximate CF another original observation that has the same labels for concepts as the genuine CF. 

The main difficulty in implementing this method is that CEBaB is sparse for the labels of the concepts on which we want to intervene, which prevents observations from being sampled directly. To alleviate this problem, following \citep{abraham2022cebab}, we trained an aspect-level classifier to predict all the concept labels for an observation. The labels predicted by this aspect-level classifier are then used to build sets of original observations that are assumed to have the same aspect labels. Random sampling of approximate CFs is performed from these sets. This method does not guarantee that there is at least one observation in the set of original observations whose predicted concept labels are identical to those of the counterfactual we intend to replace.

Our aspect-level classifier is composed of several classifiers, one per concept, trained independently to predict the value of each concept from the representation $x(s)$ of an observation $s$. We opted for simplicity by training in parallel MLP classifiers with a hidden layer of size 128 for each concept treated as ternary (the labels to predict are Positive, Negative or Unknown). This aspect-level classifier differs from the one described in \citep{abraham2022cebab} and performs less well, with an average accuracy on concepts of 68\% (which is well above random for 3-way classification).

\clearpage
\section{Additional training details}
\label{sec:training_details_complement}

All linear predictors, MLPs and regressions are based on architectures from the {\tt scikit-learn} library\footnote{\url{https://scikit-learn.org/stable/}}.
To encode the observations or finetune Bert models, we rely on the HuggingFace {\tt transformers} library\footnote{\url{https://github.com/huggingface/transformers}}. 

\subsection*{Training details for EEEC+ (sections \ref{sec:direct_comp} and \ref{sec:treatment_effect})}

Each observation is represented by the last hidden state of a frozen non-finetuned Bert (bert-base-uncased) \citep{devlin2019BERT} over the [CLS] token. The feature dimension is 768. 

Linear regressions via SGD to compute $\mu^\parallel$ for each gender value have been trained with the following parameters: learning rate is set to $1e-3$ and the strength of the $L^2$ regularization is set to $5e-2$.

Linear regressions via SGD to compute $\mu^\parallel$ for each race value have been trained with the following parameters: learning rate is set to $1e-3$ and the strength of the $L^2$ regularization is set to $5e-4$.

Linear predictor $\widehat{Y}$ (resp. $\widehat{Z}$) has been trained as one-vs-all logistic regression with $L^2$-regularization. The strength of the $L^2$ regularization is set to $1e-4$ (resp. $1e-5$).

\subsection*{Training details for CEBaB (section \ref{sec:treatment_effect_realistic})}

Each observation is represented by the last hidden state of a frozen previously finetuned Bert (bert-base-uncased) (Devlin
et al., 2019) over the [CLS] token. The feature dimension is 768. 

Bert's prior finetuning was performed on the 5-way sentiment rating prediction task. We use a maximum sequence length of 128 with a batch size of 32 and a learning rate of $5e-5$. The number of epochs for finetuning is 10.

Linear regressions via SGD to compute $\mu^\parallel$ for each aspect value and each setting have been trained with the following parameters: learning rate is set to $1e-2$ and the strength of the $L^2$ regularization is set to $1e-4$.

Linear predictor $\widehat{Y}$ has been trained as one-vs-all logistic regression with $L^2$-regularization. The strength of the $L^2$ regularization is set to $1e-5$.

\subsection*{Training details for BiasInBios (sections \ref{sec:explaining} and appendix \ref{sec:fairness})}

Each observation is represented by the last hidden state of a frozen non-finetuned Bert (bert-base-uncased) \citep{devlin2019BERT} over the [CLS] token. The feature dimension is 768. 

Linear regressions via SGD to compute $\mu^\parallel$ for each gender value have been trained with the following parameters: learning rate is set to $1e-3$ and the strength of the $L^2$ regularization is set to $5e-2$.

Linear predictor $\widehat{Y}$ (resp. $\widehat{Z}$) has been trained as one-vs-all logistic regression with $L^2$-regularization. The strength of the $L^2$ regularization is set to $1e-4$ (resp. $1e-5$).

\subsection*{Training details for GloVe dataset (appendix \ref{sec:explicit_CFs})}

Each word is represented by its GloVe embedding\footnote{\url{https://nlp.stanford.edu/projects/glove/}}. The feature dimension is 300. 

Linear regressions via SGD to compute $\mu^\parallel$ for each gender value have been trained with the following parameters: learning rate is set to $1e-3$ and the strength of the $L^2$ regularization is set to $5e-2$.

\end{document}